\documentclass[twoside]{article}

\usepackage[preprint]{aistats2027}

\usepackage[round]{natbib}

\usepackage{amsmath,amssymb,amsthm}
\usepackage{mathtools}
\mathtoolsset{showonlyrefs}

\usepackage{bm}         %
\usepackage{caption}    %
\usepackage{enumitem}   %
\usepackage{hyperref}
\usepackage{subcaption} %
\usepackage{url}        %
\usepackage[table,xcdraw,dvipsnames]{xcolor}    %

\newif\ifrevision
\revisiontrue
\ifrevision

\else

\fi

\newcommand{\jover}[2]{\overset{(#1)}{#2}}

\newcommand{\lrp}[1]{\left( #1 \right)}                     %
\newcommand{\lrb}[1]{\left[ #1 \right]}                     %
\newcommand{\lrc}[1]{\left\{ #1 \right\}}                   %
\newcommand{\lra}[1]{\left| #1 \right|}                     %
\newcommand{\lrn}[1]{\left\lVert #1 \right\rVert}           %
\newcommand{\lrang}[1]{\left\langle #1 \right\rangle}                  %

\newcommand{\ip}[2]{\left\langle{#1}\right\rangle_{#2}} %
\newcommand{\norm}[2]{\left\|{#1}\right\|_{#2}} %
\renewcommand{\d}{\mathrm d}
\renewcommand{\H}{\mathcal H}

\newcommand{\R}{\mathbb{R}}                                 %
\newcommand{\N}{\mathbb{N}}                               %
\newcommand{\E}{\mathbb{E}}                      %
\renewcommand{\S}{\mathbb{S}}             %
\newcommand{\m}[1]{\mathcal{#1}}            %

\NewCommandCopy{\textbaraccent}{\b}   %
\ExplSyntaxOn
\RenewDocumentCommand{\b}{m}
  {
    \mode_if_math:TF
      {
        \tl_if_head_is_N_type:nTF {#1}
          { \exp_args:Ne \__mybold:Nn { \tl_head:n {#1} } {#1} }
          { \bm{#1} }
      }
      { \textbaraccent{#1} }
  }
\cs_new_protected:Npn \__mybold:Nn #1#2
  { \token_if_letter:NTF #1 { \mathbf{#2} } { \bm{#2} } }
\ExplSyntaxOff

\newcommand{\negquad}{\hspace{-1em}}    %
\newcommand{\sign}{\operatorname{sign}}
\newcommand{\iid}{\overset{\mathrm{iid}}{\sim}}     %
\newcommand{\aew}{\textnormal{-a.e.}}      %

\newcommand{\trace}{\operatorname{tr}}

\newcommand{\flipneg}{\mathord{\raisebox{\heightof{$\neg$}}{\scalebox{1}[-1]{$\neg$}}}}     %
\newcommand{\transpose}{\mathsf{T}}

\newcommand{\modelYx}{{Q_{Y|x}}}
\newcommand{\modelYX}{{Q_{Y|X}}}
\newcommand{\model}{{Q_{Y|\cdot}}}

\newcommand{\refdistrYx}{{P_{Y|X=x}}}
\newcommand{\refdistrYX}{{P_{Y|X}}}
\newcommand{\refdistr}{P}
\newcommand{\refdistrX}{{P_{X}}}

\newtheorem{theorem}{Theorem}          %
\newtheorem{corollary}[theorem]{Corollary}  %
\newtheorem{lemma}[theorem]{Lemma}          %

\newtheorem{definition}[theorem]{Definition} 
\newtheorem{assumption}[theorem]{Assumption}

\theoremstyle{remark}                           %
\newtheorem{remark}[theorem]{Remark}

\setenumerate{labelindent=0em,leftmargin=2em,topsep=0cm,partopsep=2mm,parsep=0cm,itemsep=1mm}
\setitemize{labelindent=0em,leftmargin=1.3em,topsep=0cm,partopsep=0cm,parsep=0cm,itemsep=1mm}

\theoremstyle{plain}
\newtheorem{theoremA}{Theorem}[section]

\begin{document}

\twocolumn[

\title{Conditional Kernel Stein Discrepancy}

\aistatstitle{Conditional Kernel Stein Discrepancy}

\aistatsauthor{Federico Matteucci \And Florian Kalinke}

\aistatsaddress{Karlsruhe Institute of Technology \And Karlsruhe Institute of Technology}
]

\begin{abstract}%
Kernel Stein discrepancies (KSDs) provide a versatile tool for comparing distributions. 
One of their main applications is in quantifying the goodness-of-fit (GoF) between a data-generating distribution and a prescribed target distribution. 
In this work, we study the related problem of conditional GoF quantification: given only a (possibly non-normalized) conditional target model, without information on the distribution of its covariates, and samples from a joint distribution, the goal is to assess how well 
the conditional distribution of the samples matches the target. To tackle this setting, we present a framework that allows lifting unconditional KSDs to the conditional setting through an operator-valued kernel on the covariate space, going beyond the known Euclidean case. 
We establish that our suggested statistic vanishes if and only if the conditional model and the true conditional distribution agree for almost all covariates and deploy it to test conditional GoF on smooth manifolds and on discrete spaces. 
Our experiments on level, power, and runtime demonstrate the viability of testing on these domains using the proposed statistic.
\end{abstract}

\section{INTRODUCTION}

Given a sampling distribution $P$ and a target distribution $Q$, quantifying and testing their goodness-of-fit (GoF) is a fundamental problem in machine learning and statistics \citep{ingster03goftesting,lehmann21testing}, with a history dating back at least to \citet{kolmogorov33sulla,smirnov48gof}.
A recent approach for tackling this problem builds on kernel Stein discrepancies (KSDs; \citealp{liu16kernelized,chwialkowski16kernel,liu26probabilisticinferencelearningsteins}), which combine Stein's method \citep{stein72bound,chen21stein,anastasiou23stein} and reproducing kernel Hilbert spaces (RKHS; \citealp{aronszajn50theory,steinwart08support,paulsen16introduction}). In particular, KSDs involve a Stein operator---associated to $Q$---and consider the RKHS norm of the kernel mean embedding \citep{berlinet04reproducing,smola07hilbert,gretton12kernel,muandet17kernel} 
of $P$ into an RKHS of functions mapped with this operator.

Part of the appeal of KSDs stems from the fact that some Stein operators, such as the Langevin-Stein operator on Euclidean spaces \citep{gorham15measuring,liu16kernelized,chwialkowski16kernel,oates17control,gorham17measuring}, allow handling targets $Q$ of which the Lebesgue density is known only up to normalization, a setting arising, for instance, in Bayesian contexts or in the analysis of restricted Boltzmann machines.
Stein operators are known on multiple other domains, such as manifolds \citep{xu20directionalksd,xu21gofriemannianmanifold,barp22riemannksd,qu25theoryapplicationskernelsteins}, Hilbert spaces \citep{wynne25fourierksd}, discrete sets \citep{yang18discrete}, point-processes \citep{yang19point}, random graphs \citep{xu21gofgraph}, sequential models \citep{baum23ksd}, and time-to-event data \citep{fernandez20ksdgof}. %
Indeed, KSDs enjoy numerous successful applications, for example, in testing if samples come from a generative model \citep{jitkrittum17lineartime,schrab22efficient}, in medical contexts \citep{fernandez20ksdgof}, for handling wind data \citep{xu21gofriemannianmanifold}, in Bayesian or variational inference \citep{liu16bayesianinference,matsubara22robustbayesian,yu26kernelapproachsemiimplicitvariational}, and in geophysics \citep{hou25calibratinggeophysicalpredictionsconstrained}. 
We note that all KSDs listed so far can be realized as instances of the general KSD framework \citep{hagrass2026stein,kalinke26nystroemksdtests}.

A (presumably) different setting is that of assessing the GoF of conditional distributions \citep{andrews97conditional,zheng00conditional}, arising, for example, in model criticism \citep{stute02modelchecks,moreira03conditionallikelihoodratio}. 
For tackling conditional GoF, by building on the Langevin-Stein operator and vector-valued RKHSs \citep{schwartz64vrkhs,carmeli06vrkhsmercer,carmeli10vector}, \citet{jitkrittum20testing} introduced the kernel conditional Stein discrepancy (KCSD), allowing for model criticism also of non-normalized models. 
In turn, the KCSD has found successful applications, for example, in evaluating graph neural networks for multi-agent trajectory prediction \citep{li21multiagent}, in statistical testing under distribution shift \citep{thoma23testing}, in model calibration testing \citep{glaserFastScalableScorebased2023}, and
in prediction explanation \citep{sarvmaili25predictionexplanation}.
However, apart from specialized extensions, for example, \citet{glaser25computationalbiology} in computational biology, KCSDs are limited to Euclidean spaces and do not feature the broad applicability enjoyed by KSDs.

In this work, we establish that the general KSD framework encapsulates quantifying conditional GoF; we dub the construction conditional kernel Stein discrepancy (CKSD) to emphasize that it \emph{is} a KSD.
In particular, our main contributions are as follows.
\begin{itemize}
    \item The CKSD statistic allows lifting known unconditional KSDs to the conditional setting, expanding the domain of applicability of KSD-based conditional GoF tests beyond the Euclidean case.
    \item Our main results (Theorem~\ref{thm:mainresult} and Corollary~\ref{corr:cksd-is-characterizing}) establish that the proposed CKSD statistic is zero if and only if the conditional model correctly captures the data-generating distribution.
    \item We illustrate CKSD on smooth manifolds and on discrete sets. Our experiments regarding level, power, and runtime show its successful deployment on these domains.\footnote{\label{fn:code}Our code is available at \url{https://github.com/DrCohomology/CKSD}.}
\end{itemize}

\textbf{Outline}\; Section~\ref{sec:background} introduces our notation and recalls the general KSD framework.
In Section~\ref{sec:CKSD}, we discuss the conditional GoF problem, our CKSD statistic, and properties of its estimator. 
Section~\ref{sec:example-applications} instantiates CKSD on smooth manifolds and on discrete spaces.
Experiments are in Section~\ref{sec:experiments}.

\section{BACKGROUND}
\label{sec:background}

In this section, we cover the notation needed for our construction and recall the general KSD framework. %
We collect the application-specific notation in the respective subsections of Section~\ref{sec:example-applications}.

\subsection{Notation} 
\label{sec:notations}

Our notation is as follows.

\textbf{General}\;
The set of positive integers is $\mathbb N \coloneq \{1,2,\ldots\}$. We write $\bm 0_n, \bm 1_n \in \R^n$ for the vectors of $n$ zeros and $n$ ones, respectively. For a matrix $\b A\in \R^{n\times m}$, $\b A^\transpose \in \R^{m\times n}$ denotes the transpose. 
If $\b A =[a_{ij}]_{i,j=1}^n \in\R^{n\times n}$ ($n=m$), $\trace(\b A) \coloneq \sum_{i=1}^n a_{ii}$ is its trace, and $\det(\b A)$ is its determinant. 
For $\b A = [a_{ij}]_{i,j=1}^{n,m}, \b B =[b_{ij}]_{i,j=1}^{n,m}  \in \R^{n\times m}$, we write their Hadamard product as $\b A\circ\b B \coloneq [a_{ij}b_{ij}]_{i,j=1}^{n,m} \in \R^{n\times m}$. For sets $\m X$ and $\m Y$, $\m Y^{\m X}$ is the set of all $\m X \to \m Y$ functions; $f(x)\equiv y$ (with $f\in \m Y^{\m X}$ and $y\in\m Y$ fixed) emphasizes that $f(x) = y$ for all $x\in \m X$. $O$ is used for big O notation.

\textbf{Probability}\;
Let $(\m X, \tau_{\m X})$ be a separable topological space. We write $\m M_1^+(\m X)$ for the set of probability measures on $\m X$, understood w.r.t.\ the measurable space $(\m X, \Sigma_{\m X})$, where $\Sigma_{\m X}$ denotes the Borel $\sigma$-algebra induced by $(\m X, \tau_{\m X})$, unless otherwise specified. For $x\in\m X$, the Dirac measure is written as $\delta_x$, that is, $\delta_x(A) = 1$ if $x\in A$ and $0$ otherwise ($A\in \Sigma_{\m X}$). Given another separable topological space $(\m Y, \tau_{\m Y})$, a conditional model $\model$ on $\m X$ is a probability kernel $\m X \to \m M_1^+(\m Y)$. With $Q_X \in \m M_1^+(\m X)$, $Q_X\otimes \model \in \m M_1^+(\m X \times \m Y)$ (taken w.r.t.\ the product $\sigma$-algebra $\Sigma_{\m X}\otimes \Sigma_{\m Y}$) means the measure $A\times B \mapsto \int_{A}\modelYx(B)\d Q_X(x)$. For a random variable $X$ taking values in $\m X$, we define the random measure $\modelYX = \model \circ X$. 
Let $(\m H,\langle\cdot,\cdot\rangle_{\m H})$ be a real separable Hilbert space and $P\in \m M_1^+(\m X)$. $L^2(\m X, P; \m H)$ denotes the Hilbert space of (equivalence classes of) measurable $f \in \m H^{\m X}$ such that $\int_{\m X} \norm{f(x)}{\m H}^2\d P(x) < \infty$. For $f \in \m H^{\m X}$ measurable, $\E_{X\sim P}[f(X)] \coloneq \int_{\m X} f(x)\d P(x)$, where the integral is meant in Bochner's sense; the integral exists if $\int_{\m X}\norm{f(x)}{\m H}\d P(x) < \infty$. For a real-valued random variable $X\sim P\in\m M_1^+(\R)$, $\norm{X}{\psi_1} \coloneq \inf\{ t > 0 : \E_{X\sim P}[\exp(|X|/t)]\le 2\}$ denotes the $1$-Orlicz-norm; if $\norm{X}{\psi_1} < \infty$, $X$ is called sub-exponential. For a positive real-valued sequence $(a_n)_n$ and a sequence of real-valued random variables $(X_n)_n$, $X_n = O_P(a_n)$ means that $X_n/a_n$ is bounded in probability.

\textbf{(Reproducing Kernel) Hilbert Spaces}\;
Let $(\m G,\langle \cdot,\cdot\rangle_{\m G})$ and $(\m H,\langle \cdot,\cdot\rangle_{\m H})$ be real separable Hilbert spaces.
For $d \in \N$, $\m H^d$ is the product Hilbert space of $d$ copies of $\m H$ endowed with the inner product $\lrang{\b h, \b h'}_{\m H^d} \coloneqq \sum_{i=1}^d \lrang{h_i, h_i'}_{\m H}$, where $\b h = (h_1, \dots, h_d)^\transpose$, $\b h'= (h'_1, \dots, h'_d)^\transpose \in \m H^d$.
A linear operator $A : \m G \to \m H$ is called bounded if $\norm{A}{\mathrm{op}} \coloneq \sup_{\norm{g}{\m G}=1}\norm{Ag}{\m H} < \infty$, $\m B(\m G, \m H)$ denotes the set of bounded $\m G \to \m H$ linear operators, and we write $I_{\m G}$ for the identity on $\m G$. For $A\in \m B(\m G, \m H)$, we write the adjoint as $A^*$; it is the unique $B \in \m B(\m H, \m G)$ satisfying $\langle Ag, h\rangle_{\m H} = \langle g, Bh\rangle_{\m G}$ for all $g\in \m G$, $h\in \m H$.
Let $(\m X, \tau_{\m X})$ be a locally compact second-countable topological space. An $\m H$-valued reproducing kernel Hilbert space ($\m H$-RKHS) on $\m X$ is a Hilbert space $(\m H_K, \langle \cdot,\cdot \rangle_{\m H_K})$ such that (i) $\m H_K \subseteq \m H^{\m X}$ and (ii) for all $x\in \m X$ there exists a $C_x \ge 0$ such that $\norm{h(x)}{\m H} \le C_x\norm{h}{\m H_K}$ for all $h \in \m H_K$; in this case, for any $x\in \m X$, the evaluation map $\operatorname{ev}_x :\m H_K \to \m H$, $h\mapsto h(x)$ is a bounded linear operator, and the $\m H$-reproducing kernel ($\m H$-kernel) associated to $\m H_K$ is defined as $K : \m X \times \m X \to \m B(\m H, \m H)$, $(x_1,x_2) \mapsto \operatorname{ev}_{x_1}\operatorname{ev}_{x_2}^*$. Conversely, if $(\m G, \langle \cdot,\cdot \rangle_{\m G})$ is any real separable Hilbert space and $\gamma: \m X \to \m B(\m H, \m G)$, then there exists an $\m H$-RKHS $\m H_K$ of which $K(x_1,x_2) = \gamma_{x_1}^*\gamma_{x_2}$ (with $x_1,x_2\in \m X$) is the $\m H$-kernel. Let $C_0(\m X, \m H) \subseteq \m H^{\m X}$ be the space of continuous $\m X \to \m H$ functions vanishing at infinity. An $\m H$-kernel $K$ is called $C_0$ if $\m H_K \subseteq C_0(\m X, \m H)$; a $C_0$ $\m H$-kernel is called $C_0$-universal if $\m H_K$ is dense in $L^2(\m X, P; \m H)$ for any $P \in \m M_1^+(\m X)$. If $\m H = \R$ in the definitions of $\m H$-RKHS and $\m H$-kernel, by using that $\m B(\R,\R)  \cong \R$, one recovers scalar-valued RKHSs; we omit the `$\R$-' in this case. Let $k :\m X \times \m X \to \R$ be a measurable kernel with separable RKHS $\m H_k$. For $x_1,x_2\in \m X$, the map $x_1 \mapsto k(x_1,x_2)$ is written as $k(\cdot,x_2)$; $x_2 \mapsto k(\cdot, x_2)$ is called canonical feature map. The kernel mean embedding of $P\in \m M_1^+\m X)$ relative to $k$ is defined, if $\E_{X\sim P}\!\norm{k(\cdot,X)}{\m H_k} < \infty$, as 
\begin{align}
    \mu_k(P) \coloneq \E_{X\sim P}[k(\cdot,X)] \in \m H_k; \label{eq:def-kme}
\end{align}
the condition is guaranteed if $\sup_{x\in\m X}\sqrt{k(x,x)} < \infty$, in which case $k$ is called bounded. For $P,Q\in\m M_1^+(\m X)$ such that $\mu_k(P)$ and $\mu_k(Q)$ exist, we define the maximum mean discrepancy (MMD) as
\begin{align}
    \operatorname{MMD}_k(P,Q) \coloneq \norm{\mu_k(P)-\mu_k(Q)}{\m H_k}.
\end{align}

\subsection{Kernel Stein Discrepancy} \label{sec:general-ksd}
We first recall Stein operators, Stein discrepancies, and, in particular, their kernel variant.

A Stein operator $T_Q$ associated to a target probability measure $Q \in \m M_1^+(\m Z)$ on a topological space $(\m Z,\tau_{\m Z})$ is a linear operator acting on a class of functions $\m F$ on $\m Z$; it characterizes $Q$ relative to $\m F$ if, for any $P\in \m M_1^+(\m Z)$,  
\begin{align}
    Q &= P & \text{iff} && \E_{Z\sim P}[(T_Qf)(Z)] = 0,
\end{align}
for all $f \in \m F$.
The operator $T_Q$ satisfies the Stein identity (relative to $\m F$) if $\E_{Z\sim Q}[(T_Qf)(Z)] = 0$ for all $f\in \m F$. 
Given $T_Q$ and $\m F_0 \subseteq \m F$, one may define a Stein discrepancy 
\begin{equation}
\label{eq:general-stein-discrepancy}
    D(\m F_0, T_Q, P) \coloneq \sup_{f\in\m F_0}\| \E_{Z\sim P} [(T_Qf)(Z)]\|^*,
\end{equation}
where $\|\cdot\|^*$ is a suitable norm and $P\in\m M_1^+(\m Z)$; if $T_Q$ is characterizing relative to $\m F_0$, then $D(\m F_0,T_Q,P) = 0$ iff $P=Q$ \citep{anastasiou23stein}.

Choosing $\m F$ as a Hilbert space $(\H,\ip{\cdot,\cdot}{\H})$ of functions on $\m Z$ and a map $\psi_Q : \m Z \to \H$ that is Bochner-integrable\footnote{Defining KSDs in terms of Pettis integrals is also possible \citep{barp24targeted}; for simplicity, we only consider Bochner integrals throughout this manuscript.} and satisfies
\begin{equation}
    \label{eq:stein_featuremap}
    \E_{Z\sim Q} [\psi_Q(Z)] = 0_{\m H},
\end{equation}
one obtains a Stein operator on $\H$ associated to $Q$ by %
\begin{equation}
\label{eq:stein_op}
    (T_Qh)(z) \coloneqq \ip{h, \psi_Q(z)}{\H},
\end{equation}
which satisfies, for any $h\in \H$,
\begin{align}
    \E_{Z\sim  Q}[(T_Qh)(Z)] \overset{\eqref{eq:stein_op}}{=} \E_{Z\sim  Q}\ip{h,\psi_Q(Z)}{\H} = 0, \label{eq:general-ksd-Stein-identity}
\end{align}
where the last equality follows by exchanging the expectation with the inner product \citep[(A.32)]{steinwart08support} and \eqref{eq:stein_featuremap}. In other words, $T_Q$ satisfies the Stein identity relative to $\H$.\footnote{We refer to \citet[Proposition~1]{kalinke26nystroemksdtests} regarding further properties of $T_Q$.}

Now, specializing \eqref{eq:general-stein-discrepancy} by using the Stein operator \eqref{eq:stein_op}, choosing $\m F_0 = \{h \in \H : \norm{h}{\H} \le 1\}$, 
and using as `the suitable norm' the standard absolute value on $\R$ (\mbox{$\|\cdot\|^* = | \cdot |$}) yields the kernel Stein discrepancy \citep[(1)]{hagrass2026stein}
\begin{align}
    D_Q(P) \coloneq D(\m F_0,T_Q,P) = \norm{\mu_{k_Q}(P)}{\H_{k_Q}}, \label{eq:ksd}
\end{align}
where $\mu_{k_Q}$ is as in \eqref{eq:def-kme} with $k_Q : \m Z\times \m Z  \to \R$  a kernel on $\m Z$ given by 
\begin{align}
    (z, z^\prime) \mapsto \ip{\psi_{Q}(z), \psi_Q(z^\prime)}{\m H}, \label{eq:general-ksd-stein-kernel}
\end{align}
having associated RKHS $\H_{k_Q}$; $D_Q(P)$ exists if $\E_{Z\sim P}\!\norm{k_Q(\cdot,Z)}{\m H_{k_Q}} < \infty$. 
We note that both the feature map $\psi_Q(z) \in \m H$ and the canonical feature map $k_Q(\cdot, z) \in \m H_{k_Q}$ induce the kernel $k_Q$, despite mapping to different Hilbert spaces.

\section{CONDITIONAL KERNEL STEIN DISCREPANCY}
\label{sec:CKSD}

In this section, we recall the conditional GoF testing problem, %
define our statistic for tackling it, %
and provide its main statistical and computational
guarantees. %

\subsection{Problem Statement} \label{sec:problem-statement}

We start with the following assumption.
\begin{assumption} \label{ass:well-defined-problem}
     $(\m X,\tau_{\m X})$ is a topological space and $(\m Y,\tau_{\m Y})$ is a Polish space.
\end{assumption}

\begin{remark} \label{remark:ass:well-defined-problem} Assumption~\ref{ass:well-defined-problem} guarantees the existence of regular conditional probability measures \citep[Theorem~8.5]{kallenberg21foundations}.
\end{remark}

Given Assumption~\ref{ass:well-defined-problem}, a conditional model $\model : \m X \to \mathcal M_1^+(\m Y)$, and samples $(X_i, Y_i)_{i=1}^n \iid P \in \m M_1^+(\m X\times \m Y)$, we test
\begin{align}
\begin{split}
    &H_0: \modelYX \overset{\refdistrX\aew }{=} \refdistrYX \quad \text{vs.} \\ 
    &H_1 : \exists E \in \Sigma_{\m X} \text{ s.t.}\, 
        \begin{cases}
             \refdistrX(E) > 0,\\
             \forall x \in E, \modelYx \neq \refdistrYx,
        \end{cases}
\end{split}
\label{eq:gof_null}
\end{align}
where $P_X$ is the marginal distribution of $P$ on $\m X$ and $X\sim P_X$.
Notice that the data-generating distribution $\refdistr$, also called sampling distribution in the following, is unknown and only observed through the sample $(X_i,Y_i)_{i=1}^n$; moreover, the model $\model$ does not specify the marginal distribution of $X$. 
Indeed, this is the main difference between testing a joint model for $(X,Y)$ and a conditional model.

To tackle this problem, we design an appropriate statistic in the next section.

\subsection{Proposed Statistic}
\label{sec:cksd-statistic}

In the following, we impose the assumption below, expanding Assumption~\ref{ass:well-defined-problem}.

\begin{assumption}\label{ass:well-defined-RKHS}
    (i) $(\m X,\tau_{\m X})$ is a locally compact second-countable topological space, $(\m Y,\tau_{\m Y})$ is a Polish space, and $\m H$ and 
$\m G \subseteq \m H^{\m X}$ are real separable Hilbert spaces.\footnote{\label{fn:g-identification} We identify $\m G$ with the corresponding subset of $\m H^{\m X \times \m Y}$ consisting of $\m Y$-independent functions.} (ii) There exists a $P$-Bochner integrable family of conditional Stein feature maps $\lrc{\psi_\modelYx}_{x\in \m X} \subseteq \m H^{\m Y}$, satisfying, for $\refdistrX\aew\ x\in\m X$,
\begin{align}
    \modelYx &= \refdistrYx \text{ iff } \ \E_{Y\sim \refdistrYx}[\psi_\modelYx(Y)] = 0_{\m H}.\quad\label{eq:conditional-stein-characterizing}
\end{align}
\end{assumption}

\begin{remark}\label{remark:ass:well-defined-RKHS}~
\begin{enumerate}[label=(\alph*)]
    \item \textbf{Topological requirements.} The second-countability and local compactness of $\m X$ ensure that the $\m H$-RKHS associated to a $C_0$-universal kernel $K : \m X  \times \m X \to \m B(\m H,\m H)$ is dense in $L^2(\m X, P_X; \m H)$ by \citet[Theorem~1]{carmeli10vector}, which will be at the root of our main result (Theorem~\ref{thm:mainresult}). 
    \item \textbf{Conditional Stein feature maps.} We show in Section~\ref{sec:example-applications} that existing unconditional kernel Stein feature maps can be leveraged to construct $\psi_\modelYx$-s satisfying \eqref{eq:conditional-stein-characterizing}. Moreover, we recall that one of the key benefits of using KSDs for GoF testing is their ability to also handle non-normalized target distributions. If the unconditional Stein feature maps are agnostic to normalization, the induced conditional ones are so as well. 
\end{enumerate}

\end{remark}

Given Assumption~\ref{ass:well-defined-RKHS}, let $\lrc{\gamma_x}_{x\in\m X} \subseteq \m B(\m H, \m G)$ be a family of uniformly bounded linear operators and define
\begin{align}
    \Psi_Q: \m X \times \m Y \to \m G,\quad (x,y) \mapsto \gamma_x \psi_\modelYx(y). \label{eq:def-psi-Q}
\end{align}
With $\m Z \mapsto \m X \times \m Y$, $\psi_Q \mapsto \Psi_Q$, and  $\m H \mapsto \m G$ in \eqref{eq:stein_op},\textsuperscript{\ref{fn:g-identification}} we get the Stein operator $T_Q$ on $\m G$, that is, for $g\in \m G$, 
\begin{align}
    (T_Qg)(x,y) = \ip{g,\Psi_Q(x,y)}{\m G}.
\end{align}
Importantly, these definitions put us into the context of the general KSD framework detailed in Section~\ref{sec:general-ksd} and exhibit two key properties:
(i) $T_Q$ does not depend on the unknown $\refdistrX$;
(ii) $T_Q$ satisfies a Stein identity relative to $\m G$ for 
\begin{align}
    Q=Q_X\otimes \model \in \m M_1^+(\m X \times \m Y), \label{eq:q-is-qx-times-model}
\end{align}
where $Q_X \in \m M_1^+(\m X)$ is arbitrary, as we establish next. 
Indeed, we obtain
\begin{align}
    \MoveEqLeft\E_{(X,Y)\sim\refdistr}[ \Psi_Q(X, Y)] \\
        &\jover a= \E_{X \sim \refdistrX} [\E_{Y\sim \refdistrYX} [\gamma_X \psi_\modelYX(Y)]] \\
        &\jover b= \E_{X\sim\refdistrX}[ \gamma_X \E_{Y\sim\refdistrYX} [\psi_\modelYX(Y)]], \label{eq:exp-of-psi-Q}
\end{align}
where in (a) we disintegrate the measure $P=P_X\otimes P_{Y\mid X}$ (valid by Remark~\ref{remark:ass:well-defined-problem}) and use \eqref{eq:def-psi-Q}; in (b), we exchange the integral with the bounded linear operator, valid by \citet[(A.32)]{steinwart08support}.
As a direct consequence, using the linearity of the $\gamma_x$-s and \eqref{eq:conditional-stein-characterizing}, we have
\begin{align}
    \MoveEqLeft\E_{Y\sim \refdistrYX} [\psi_\modelYX(Y)] \overset{\refdistrX\aew }{=} 0_{\m H}\\
    &\implies \E_{(X,Y)\sim\refdistr} [\Psi_Q(X, Y)] = 0_{\m G}.   \label{eq:sufficient-Stein-condition}
\end{align}
Hence, as in \eqref{eq:general-ksd-Stein-identity}, for any $g\in\m G$, we get that $\E_{(X,Y)\sim Q}[(T_Qg)(X,Y)] = 0$  (with $Q$ as in \eqref{eq:q-is-qx-times-model}), that is, $T_Q$ inherits a Stein identity from the characterizing property of the $\psi_{\modelYx}$-s. 

We establish the converse to \eqref{eq:sufficient-Stein-condition} in the following result, proved in Appendix~\ref{sec:proof-mainresult}. The result can be seen as an extension of \citet[Theorem 1]{jitkrittum20testing} to general spaces. In particular, we do not require the measures to be absolutely continuous w.r.t.\ the Lebesgue measure on $\R^d$.

\begin{theorem}[$\Psi_Q$ is characterizing]
\label{thm:mainresult}
    Given Assumption~\ref{ass:well-defined-RKHS}, let $K: \m X \times \m X \to \m B\lrp{\m H, \m H}$ with $K(x, x') = \gamma_x^* \gamma_{x'}$ be a $C_0$-universal $\m H$-kernel, and $\Psi_Q$ as in \eqref{eq:def-psi-Q}. Further assume that $\E_{P_{Y\mid X= \cdot}} [\psi_\model(Y)] \in L^2(\m X,P_X;\m H)$.
    Then $
        \E_{(X,Y)\sim \refdistr}[ \Psi_Q(X, Y)] = 0_{\m G}$
        iff
        $\E_{Y\sim\refdistrYX} [\psi_\modelYX(Y)] \overset{\refdistrX\aew}{=} 0_{\m H}$.   
\end{theorem}

With the above result in place, we next present our test statistic and its validity. 
Performing the substitutions stated underneath \eqref{eq:def-psi-Q} in \eqref{eq:general-ksd-stein-kernel}, we get
\begin{align}
    \MoveEqLeft k_Q((x, y), (x^\prime, y^\prime))  \overset{\eqref{eq:general-ksd-stein-kernel}}{=} \lrang{\Psi_Q(x, y), \Psi_Q(x', y')}_{\m G}\\
        &\overset{\eqref{eq:def-psi-Q}}{=} \lrang{\gamma_x \psi_\modelYx(y), \gamma_{x'}\psi_{Q_{Y|x'}}(y')}_{\m G}\\
        &\jover a= \lrang{ \psi_\modelYx(y),\gamma_x^* \gamma_{x'} \psi_{Q_{Y|x'}}(y')}_{\m H}\\
        &\jover b= \lrang{\psi_\modelYx(y), K(x, x') \psi_{Q_{Y|x'}}(y')}_{\m H}, \label{eq:def-k_Q}
\end{align}
where in (a) we consider the adjoint of $\gamma_{x}$; in (b) 
we use $K(x, x') = \gamma_x^*\gamma_{x'}$, as in Theorem~\ref{thm:mainresult}.

Finally, collecting our previous derivations and combining them with \eqref{eq:ksd} motivates our following conditional kernel Stein discrepancy (CKSD) statistic.
\begin{definition}[CKSD Statistic] \label{def:cksd-statistic}
Given Assumption~\ref{ass:well-defined-RKHS} and $k_Q$ as in \eqref{eq:def-k_Q}, define 
\begin{align}
    \MoveEqLeft\m P = \big\{P\in \m M_1^+(\m X\times\m Y) : \\
    &\quad\E_{(X,Y)\sim P} \norm{k_Q(\cdot,(X,Y))}{\m H_{k_Q}}< \infty\big\}.\notag
\end{align}
For $P \in \m P$, let
\begin{align}
    D_Q(\refdistr) = \lrn{\mu_{k_Q}(\refdistr)}_{\m H_{k_Q}}. \label{eq:cksd-def}
\end{align}
\end{definition}

\begin{remark}
    Restricting the definition of $D_Q$ to $P\in\m P$ ensures that the population quantity is well defined.
\end{remark}
 
The following corollary (proved in Appendix~\ref{sec:proof-cksd-is-characterizing}) justifies leveraging the standard KSD framework to test for \eqref{eq:gof_null}.

\begin{corollary}[Validity of $D_Q(P)$] \label{corr:cksd-is-characterizing}
    Under the assumptions of Theorem~\ref{thm:mainresult}, it holds for $P \in \m P$ that
    \begin{align}
        D_Q(\refdistr) &= 0 && \text{iff} & \modelYX \overset{\refdistrX\aew}{=} \refdistrYX.
    \end{align}
\end{corollary}

Given a sample $(X_i,Y_i)_{i=1}^n\iid P \in \mathcal P$, a V-statistic-based estimator of $D_Q(\refdistr)$ is obtained by replacing $\refdistr$ with the empirical measure $\hat\refdistr_n = \frac1n\sum_{i=1}^n\delta_{(X_i,Y_i)}$ and squaring, captured in the following lemma, which we derive in Appendix~\ref{sec:proof-estimator}.

\begin{lemma}[V-statistic-based CKSD Estimator] \label{lemma:estimator} Given the setting of Definition~\ref{def:cksd-statistic} and a sample $(X_i,Y_i)_{i=1}^n\iid P$,
\begin{align}
    D_Q^2(\hat \refdistr_n) = \frac{1}{n^2}\bm 1^\transpose_n\b K_{n,n}\bm 1_n, \label{eq:v-stat-estimator}
\end{align}
    with $\b K_{n,n}=[k_Q((X_i,Y_i),(X_j,Y_j))]_{i,j=1}^n \in \R^{n\times n}$. Moreover, if $k : \m X \times \m X \to \R$ is a kernel and $K(x,x') = k(x,x')I_{\m H}$, then
    \begin{align}
       D_Q^2(\hat \refdistr_n) &= \frac{1}{n^2}\bm 1^\transpose_n(\b L_{n,n} \circ \b G_{n,n})\bm 1_n, \label{eq:v-stat-estimator-simplified}%
    \end{align}
    where $\b L_{n,n}=[k(X_i,X_j)]_{i,j=1}^n \in \R^{n\times n}$ and $\b G_{n,n}=[\langle \psi_{Q_{Y\mid X_i}}(Y_i),\psi_{Q_{Y\mid X_j}}(Y_j)\rangle_{\m H}]_{i,j=1}^n \in \R^{n\times n}$.
\end{lemma}

We next detail the properties of the V-statistic-based CKSD estimator.  

\subsection{Additional Guarantees} \label{sec:guarantees}

This section collects additional guarantees on the estimator \eqref{eq:v-stat-estimator}. In particular, the next result (proved in Appendix~\ref{sec:proof-of-consistency}) establishes its consistency; we comment on its null distribution, runtime complexity, and a possible acceleration after elaborating the result.

\textbf{Consistency}\;
The following lemma establishes the rate of \eqref{eq:v-stat-estimator}, implying its consistency. 

\begin{lemma}[$\sqrt n$-rate] \label{lemma:convergence-rate}
Let $P\in\m P$, $D_Q(P)$ as in \eqref{eq:cksd-def}, $(X_i,Y_i)_{i=1}^n \iid P$, $D_Q(\hat P_n)$ the square root of \eqref{eq:v-stat-estimator}, and $K : \m X \times \m X \to \m B(\m H,\m H)$ a $C_0$ $\m H$-kernel. If $ \norm{\norm{\psi_{\modelYX}(Y)}{\m H}}{\psi_1} < \infty$, it holds that
    \begin{align}
        D_Q(P) - D_Q(\hat P_n) = O_P(n^{-1/2}).
    \end{align}
\end{lemma}

\begin{remark}\label{remark:lemma:convergence-rate}~ 
    \begin{enumerate}[label=(\alph*)]
        \item \textbf{Sub-exponential assumption.} Classical results on the convergence of kernel mean embedding-based estimators assume bounded feature maps \citep{smola07hilbert,gretton12kernel,lopezpaz15towardslearning,tolstikhin17minimax}. However, in the context of KSDs, feature maps are often unbounded (\citealp[Example~1]{kalinke25nystromksd}; \citealp[Remark~4]{hagrass2026stein}). Imposing a sub-Gaussian \citep[Theorem~3]{kalinke25nystromksd} or a weaker sub-exponential \citep[Theorem~4]{kalinke26nystroemksdtests} assumption---as in Lemma~\ref{lemma:convergence-rate}---allows one to obtain the parametric $\sqrt n$-rate. We  note that \citet[Example~3]{kalinke25nystromksd} illustrates the analytical verification of a KSD feature map satisfying sub-Gaussianity in a simple case.
        \item \textbf{Optimal rate.} It is known \citep{cribeiroramallo25minimaxlowerboundkernel,cribeiroramallo26minimaxlowerboundskernel,gogolashvili26minimaxestimationkernelstein} that the $\sqrt n$-rate, guaranteed to hold in the setting of Lemma~\ref{lemma:convergence-rate}, is asymptotically optimal in the minimax sense.
    \end{enumerate}
\end{remark}

\textbf{Null Distribution}\;
The distribution of \eqref{eq:v-stat-estimator} under both the null and the alternative hypotheses follows from known results on V-statistics \citep{serfling80approximation}; it is similarly implied by known results on the general KSD \citep[Theorem~6]{kalinke26nystroemksdtests}. However, in general, the null distribution is analytically intractable. If $(\m X\times \m Y,d_{\m X\times \m Y})$ is a metric space and $\E_{(X,Y)\sim P}[k_Q((X,Y),(X,Y))] < \infty$, one may instead use the wild (also known as i.i.d.\ weighted) bootstrap to approximate the null distribution \citep{dehling94bootstrap,chwialkowski16kernel,kalinke26nystroemksdtests}. Indeed, given (fixed) $(x_i,y_i)_{i=1}^n \in (\m X\times \m Y)^n$, let
\begin{align}
    nB_n^2 \coloneq \frac1n\sum_{i,j=1}^nW_iW_j k_Q((x_i,y_i),(x_j,y_j)),\label{eq:bootstrap}
\end{align} 
where the $W_i$-s are Rademacher random variables, that is, they are $\pm1$ with equal probability. Under $H_0$, $nD_Q^2(\hat \refdistr_n)$ and \eqref{eq:bootstrap} have the same distributional limit, and one may approximate quantiles by numerically approximating the law of \eqref{eq:bootstrap}.

\textbf{Runtime}\;
The computation of $D_Q^2(\hat \refdistr_n)$ in \eqref{eq:v-stat-estimator} requires $O(n^2)$ evaluations of $k_Q$ for computing $\b K_{n,n}$. Summing these $n^2$ entries and normalizing the result also costs $O(n^2)$, which leads to a total runtime complexity of $O(n^2)$. The estimator \eqref{eq:v-stat-estimator-simplified} requires computing two $n\times n$ matrices, each costing $O(n^2)$, and their Hadamard product, which is also $O(n^2)$. Summing and normalizing the result is in $O(n^2)$; hence, the total cost of computing \eqref{eq:v-stat-estimator-simplified} is $O(n^2)$.

\textbf{Acceleration}\;
We note that \citet{kalinke25nystromksd,kalinke26nystroemksdtests} have shown that KSD estimators and KSD-based tests can be accelerated by the Nyström method \citep{williams01using,rudi15less,chatalic22nystrom,kalinke23nystrommhsic}. As we are within the general KSD framework, these accelerations can be applied to our proposed method, to allow tackling large-scale settings.

\section{EXAMPLE APPLICATIONS} 
\label{sec:example-applications}

In this section, we illustrate the application of our framework when $\m Y$ is either a manifold or a discrete set, with $\m X$ as in Assumption~\ref{ass:well-defined-RKHS}. 
In particular, we separately instantiate \eqref{eq:def-k_Q} with a suitably chosen family of conditional Stein feature maps $\lrc{\psi_\modelYx}_{x\in\m X}$.

Let $\ell : \m Y\times \m Y \to \R$ be a $C_0$-universal kernel, and $K: \m X \times \m X \to \m B\lrp{\m H_\ell^d, \m H_\ell^d}$ a $C_0$-universal $\m H_\ell^d$-kernel. 
We assume that $K(x, x') = k(x, x') I_{\m H_\ell^d}$, where $k : \m X \times \m X \to \R$ is a kernel.\footnote{It is known that $K$ is $C_0$-universal if and only if $k$ is \citep[Example 14]{carmeli10vector}.}

\subsection{Smooth Manifolds}
\label{sec:example-applications:manifolds}

\textbf{Additional Notation}\;
Let $\m Y$ be a $d$-dimensional smooth manifold with boundary $\partial \m Y$ and $\b \theta = \lrp{\theta_1, \dots, \theta_d}^\transpose \in \R^d$ a coordinate system in an open subset $U$ of $\m Y$, that is, $U$ is homeomorphic to an open subset of $\R^d$ via $\b \theta \mapsto y(\b \theta)$ (except on $\partial \m Y$).
For $i = 1,\dots, d$, let $\partial_{\theta_i}$ be the partial derivative in the $i$-th direction.
For a smooth $f: \m Y \to \R$, its gradient is $\b \nabla_{\b \theta}f = \lrp{\partial_{\theta_1}f, \dots, \partial_{\theta_d}f}^\transpose : \m Y \to \R^d$.
Let $J(y)$ be the volume element at $y \in \m Y$; if $\m Y$ is embedded in $\R^m$ for some $m$, that is, $y = (y_1, \dots, y_m)$,  $J(y) = \det(Dy^\transpose Dy)^{\frac12}$ where $Dy = \lrb{\partial_{\theta_j} y_i}_{i, j=1}^{m, d}$ is the Jacobian matrix of $y$.

\begin{assumption}
\label{ass:manifolds}
    (i) $\b \theta$ is a coordinate system that covers $\m Y$ a.e.
    (ii) $\ell$ is smooth.
    Either (iii) $\partial \m Y = \emptyset$ or (iii)' for any $\b f \in \m H_\ell^d$, $\b f$ vanishes on $\partial \m Y$. 
\end{assumption}

\citet{xu21gofriemannianmanifold} introduce the Stein feature map $\psi_{Q'}(y)$ associated to $Q' \in \m M_1^+(\m Y)$ with smooth probability density function (pdf) $q'$ 
\begin{align}
\label{eq:manifolds_xu_steinfeaturemap}
    \psi_{Q'}(y) = \b \nabla_{\b \theta} \ell(\cdot, y) + \ell(\cdot, y) \b \nabla_{\b \theta} (\log q'(y) J(y)),
\end{align}
and show that, given Assumption~\ref{ass:manifolds} and $P' \in \m M_1^+(\m Y)$, $\E_{Y\sim P'} \lrb{\psi_{Q'}(Y)}=0_{\m H_\ell}$ iff $P' = Q'$.

We now define the related conditional Stein feature maps.
For $\refdistrX\aew\ x\in \m X$, let $q_x$ be the smooth pdf of $\modelYx$ and, with the score function $\b s_q(x, y) \coloneqq \b \nabla_{\b \theta} (\log q_x(y) J(y))$, define the conditional Stein feature map as 
\begin{align}
    \psi_\modelYx(y) = \b \nabla_{\b \theta} \ell(\cdot, y) + \ell(\cdot, y) \b s_q(x, y) \in \m H_\ell^d, \notag
\end{align}
which satisfies Assumption~\ref{ass:well-defined-RKHS} (ii) by \citet[Theorem 1]{xu21gofriemannianmanifold}.
Finally, using 
\begin{align}
\label{eq:stein_kernel_manifold}
\MoveEqLeft\lrang{\psi_\modelYx(y), \psi_{Q_{Y|x'}}(y')}_{\m H_\ell^d}\\
    &=\sum_{i= 1}^d \partial_{\theta_i} \partial_{\theta'_i} \ell(y, y')
        + \ell(y, y')\, \b s_q(x, y)^{\transpose} \b s_q(x', y')\notag\\
        &\quad+ \b \nabla_{\b\theta} \ell(y, y')^{\transpose} \b s_q(x', y')
        + \b \nabla_{\b\theta'} \ell(y, y')^{\transpose} \b s_q(x, y) \notag
\end{align}
and that $K = kI_{\m H_\ell^d}$ gives an explicit form of $k_Q$ in \eqref{eq:def-k_Q}, allowing the computation in practice.

To conclude this section, we note that the $\psi_\modelYx$-s are agnostic to the normalization constant of the conditional pdf $q_x$.
Moreover, choosing $\m X = \R^{d_1}$, $\m Y = \R^{d_2}$, $y(\b \theta) = \b \theta$, and $J(y) \equiv 1$ in the above, we recover \citet[Proposition 2]{jitkrittum20testing}.

\subsection{Discrete Sets}
\label{sec:example-applications:discrete}

\textbf{Additional Notation}\;
Let $\m Y = \lrc{y^{\lrb{1}}, \dots, y^{\lrb{\lra{\m Y}}}}^d$ be the product of $d$ copies of a finite set with an arbitrary ordering.
Define a cyclic permutation $\neg: y^{\lrb{j}} \mapsto y^{\lrb{\pmod{j+1 \lra{\m Y}}+1}}$, whose inverse we denote with $\flipneg$.  
For $\b y = (y_1, \dots, y_d)^\transpose \in \m Y$, let $\neg_i: \b y\mapsto (y_1, \dots, \neg y_i, \dots, y_d)^\transpose$.
For a function $f: \m Y \to \R$, we define the partial difference operator $\Delta_{y_i} f(\b y) \coloneqq f(\b y) - f(\neg_i \b y)$ and write $\Delta_{\b y} f(\b y) \coloneqq \lrp{\Delta_{y_1} f(\b y), \dots, \Delta_{y_d} f(\b y)}^\transpose$.
Similarly, let $\Delta^*_{\b y}$ be the difference operator w.r.t.\ $\flipneg$.

\citet{yang18discrete} introduce the Stein feature map $\psi_{Q'}(y)$ associated to a $Q' \in \m M_1^+(\m Y)$ with probability mass function (pmf) $q'$ with full-support, in the form
\begin{align}
\label{eq:yang_discrete_steinfeaturemaps}
    \psi_{Q'}(\b y) = \frac{\Delta_{\b y} q'(\b y)}{q'(\b y)} \ell^{\otimes d}(\cdot, \b y) - \Delta_{\b y}^*\ell^{\otimes d}(\cdot, \b y),
\end{align}
where $\ell^{^{\otimes d}} = \ell \otimes \overset d{\cdots} \otimes \ell$.
They show that, given $P' \in \m M_1^+(\m Y)$, $\E_{Y\sim P'} \lrb{\psi_{Q'}(Y)}=0_{\m H_\ell}$ iff $P' = Q'$.

For $\refdistrX\aew\ x \in \m X$, assume that $\modelYx$ has pmf $q_x$ with full support. 
Next, defining the score vector as $\b s_q(x, \b y) \coloneqq {\Delta_{\b y} q_x(\b y)} / {q_x(\b y)}$, the conditional Stein feature map then takes the form
\begin{align}
\label{eq:discrete_steinfeaturemaps}
    \psi_\modelYx(\b y) = \ell^{\otimes d}(\cdot, \b y) \b s_q(x, \b y) - \Delta_{\b y}^*\ell^{\otimes d}(\cdot, \b y) \in \m H_\ell^d,
\end{align}
and satisfies Assumption~\ref{ass:well-defined-RKHS} (ii) by \citet[Lemma 8]{yang18discrete}.
Next, using that $K = kI_{\m H_\ell^d}$ and 
\begin{align}
\label{eq:stein_kernel_discrete}
\MoveEqLeft\lrang{\psi_\modelYx(y), \psi_{Q_{Y|x'}}(y')}_{\m H_\ell^d}\\
&= \trace\lrp{\Delta^*_{\b y, \b y'} \ell^{\otimes d}(\b y, \b y')} \\
&\quad+
        \ell^{\otimes d}(\b y, \b y') \b s_q(x, \b y)^\transpose \b s_q(x', \b y')\notag\\ 
&\quad- \Delta_{\b y}^* \ell^{\otimes d}(\b y, \b y')^\transpose \b s_q(x', \b y') \\
            &\quad- \Delta_{\b y'}^* \ell^{\otimes d}(\b y, \b y')^\transpose \b s_q(x, \b y), \notag
\end{align}
one obtains an explicit form of $k_Q$ in \eqref{eq:def-k_Q}.

Finally, we note that the $\psi_\modelYx$-s are agnostic to the normalization constant of the conditional pmf $q_x$.

\section{EXPERIMENTS}
\label{sec:experiments}

In this section, we perform conditional GoF tests with the proposed CKSD on manifolds (rotamers) %
and on discrete sets (permutations), %
assessing the type I error, power, and runtime.\textsuperscript{\ref{fn:code}}

As a baseline, we use a two-sample test based on the MMD with kernel $k \otimes \ell : (\m X\times \m Y) \times (\m X \times \m Y) \to \R$, defined by $((x,y),(x',y'))\mapsto k(x,x')\ell(y,y')$, with kernels $k$ and $\ell$ as in Section~\ref{sec:example-applications}.\footnote{As manifolds and discrete sets are locally compact Polish spaces,  the $C_0$-universality of $k$ and $\ell$ ensures that of $k \otimes \ell$~\citep{szabo2018characteristic}.}
The following lemma (proved in Appendix~\ref{app:proof_H0iffjointH0}) motivates using two-sample testing.
\begin{lemma}[Two-sample formulation]
\label{lemma:H0iffjointH0} Let $\refdistr = \refdistrX \otimes \refdistrYX$ and $\model$ be as in Section~\ref{sec:problem-statement}. Then
    $\refdistrYX \overset{\refdistrX\aew}{=} \modelYX $ iff $ \refdistr = \refdistrX \otimes \modelYX$.
\end{lemma}
To perform the test, we split the sample $\lrp{X_i, Y_i}_{i=1}^n \iid P$ into $\b Z_1 = \lrp{X_i, Y_i}_{i=1}^{\lfloor n/2\rfloor}$ and $\b Z_2 = \lrp{X_i, Y_i}_{i=\lfloor n/2 \rfloor+1}^{n}$.
From $\b Z_1$ we construct $\tilde{\b Z}_1 = (X_i, \tilde Y_i)_{i=1}^{\lfloor n/2\rfloor}$ by replacing $Y_i$ with $\tilde Y_i \sim Q_{Y|X_i}$ for every $i=1,\ldots, \lfloor n/2\rfloor$. 
Notice that $\tilde{\b Z}_1 \iid \refdistrX \otimes \modelYX$ and $\b Z_2 \iid \refdistrX \otimes \refdistrYX$, and that Lemma~\ref{lemma:H0iffjointH0} guarantees that $\tilde{\b Z}_1$ and $\b Z_2$ follow the same distribution iff $H_0$ in \eqref{eq:gof_null} holds.
As test statistic we therefore use the V-statistic-based estimator $\operatorname{MMD}_{k \otimes \ell}^{ 2}(\hat F_{1,n}, \hat F_{2,n})$, where $\hat F_{1,n}$ and $\hat F_{2,n}$ denote the empirical measures associated to $\tilde{\b Z}_1$ and $\b Z_2$, respectively. We use a permutation approach for simulating the null distribution.

\subsection{Rotamers} \label{sec:experiments:rotamers}

\begin{figure*}[t]
    \begin{center}
        \subfloat[$H_0$ holds]{\includegraphics[height=1.4in]{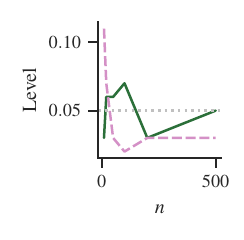}\label{fig:torus_level}}
        \subfloat[$H_1$ (global) holds]{\includegraphics[height=1.4in]{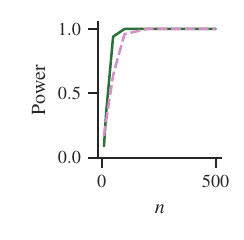}\label{fig:torus_power_Q2}}
        \subfloat[$H_1$ (local) holds]{\includegraphics[height=1.4in]{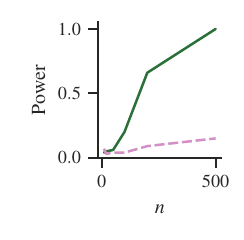}\label{fig:torus_power_Q3}}
        \subfloat[Average runtime]{\includegraphics[height=1.4in]{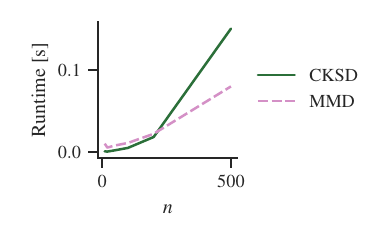}\label{fig:torus_runtime}}
    \end{center}
    \caption{
    Comparison of CKSD with an MMD-based test w.r.t.\ level, power, and runtime on rotamers. 
    }
    \label{fig:rotamers}
\end{figure*}

\begin{figure*}[t]
    \begin{center}
    \subfloat[$H_0$ holds]{\includegraphics[height=1.4in]{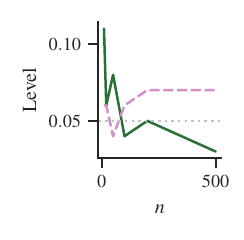}\label{fig:permutations_level}}
    \subfloat[$H_1$ (global) holds]{\includegraphics[height=1.4in]{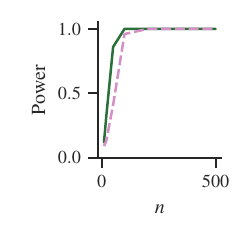}\label{fig:permutations_power_Q2}}
    \subfloat[$H_1$ (local) holds]{\includegraphics[height=1.4in]{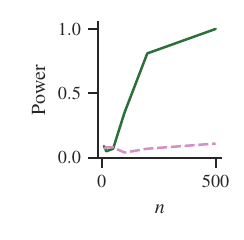}\label{fig:permutations_power_Q3}}
    \subfloat[Average runtime]{\includegraphics[height=1.4in]{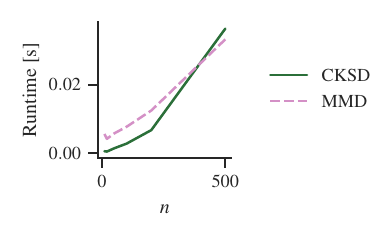}\label{fig:permutations_runtime}}
    \end{center}
    \caption{
    Comparison of CKSD with an MMD-based test w.r.t.\ level, power, and runtime on permutations.
    }
    \label{fig:permutations}
\end{figure*}

The conformation of a protein is partially described by a pair of (backbone) torsion angles, angles of rotation about chemical bonds~\citep{mardia2007rotamers}.
For proteins, a rotation about a single bond has finitely many energy minima: the torsion angles tend to concentrate around these minima, forming clusters called rotamers~\citep{shapovalov2011rotamers}.

\textbf{Formalization}\;
For simplicity, we formalize a rotamer as an element of $\m X = (-1, 1)$.
The two torsion angles $\theta_1, \theta_2 \in (-\pi, \pi]$ define a coordinate system on the torus $\m Y \coloneqq \S^1 \times \S^1$ (embedded in $\R^4$) as 
\begin{align}
\label{eq:rotamer_coordinates}
    y(\b \theta) = \lrp{\cos\theta_1, \sin \theta_1, \cos \theta_2, \sin\theta_2}^\transpose.
\end{align}

The conditional distribution of the torsion angles given a rotamer can be modeled with a cosine model, that is, a product of von Mises distributions on the two $\S^1$-s and a coupling term~\citep{mardia2007rotamers}. 
The cosine model (conditional on a rotamer $x$) has density
\begin{align}
\label{eq:cosine_model}
     f_{\b \kappa, \b \mu}(\b \theta)
        &= C \exp \big(\kappa_1\cos(\theta_1-\mu)\\
        &\negquad+\kappa_2\cos(\theta_2-\nu)
        -\kappa_3\cos(\theta_1-\mu-\theta_2+\nu) \big),
\end{align}
where $\b \kappa = (\kappa_1, \kappa_2, \kappa_3)$, $\b \mu = (\mu, \nu)$, both $\b \kappa$ and $\b \mu$ depend on $x$, and $C > 0$ is a normalizing constant.

\textbf{Stein Kernel}\;
As $\m Y$ is a smooth manifold, we apply the Stein kernel of Section~\ref{sec:example-applications:manifolds}.
In the coordinate system \eqref{eq:rotamer_coordinates}, the volume element is $J(y) \equiv 1$.

On $\m X$ we use the Gaussian kernel $k(x, x') = e^{-\gamma(x-x')^2}$ with $\gamma=1$. 
On $\m Y$, we use the product of two von Mises kernels on the individual components, that is,
\begin{align}
    \label{eq:vonMises_kernel}
    \ell(y, y') = e^{\gamma y^\transpose y'} \overset{\eqref{eq:rotamer_coordinates}}{=} e^{\gamma \lrp{\cos(\theta_1 - \theta_1') + \cos(\theta_2 - \theta_2')}},
\end{align}
with $\gamma = 1$. 
As the von Mises kernel is $C_0$-universal~\citep{park2022kernel} and $\S^1$ is locally compact Polish, $\ell$ is $C_0$-universal~\citep{szabo2018characteristic}.

\textbf{Experimental Setup}\;
We set $\refdistrX = \m U(-1, 1)$, a uniform distribution on $\m X$, and assume that $\refdistrYx$ has density $p_x \propto f_{\b \kappa, \b 0_2}$ with $\b \kappa = \lra{x} \cdot (3, 2, 1)$, ensuring that the distribution is unimodal~\citep{mardia2007rotamers} and, thus, describes one rotamer.
We show the results of the test for three models $Q^i_{Y|\cdot}$ ($i=1, 2, 3$) with respective conditional densities 
\begin{align}
    q^1_x(\b \theta) &\propto f_{\lra x \cdot (3, 2, 1), \b 0_2}(\b \theta),   \tag*{$H_0$\quad}\\
    q^2_x(\b \theta) &\propto f_{\lra {5x} \cdot (3, 2, 1), \b 0_2}(\b \theta),  \tag*{$H_1$ (global shift)\quad} \\
    q^3_x(\b \theta) &\propto 
        \begin{cases}
            q^1_x(\b \theta) \quad \text{w.p.}\ 0.8,\\
            q^2_x(\b \theta) \quad \text{w.p.}\ 0.2.
        \end{cases}
        \tag*{$H_1$ (local shift)\quad}
\end{align}
We compute the rejection rates at level $\alpha = 0.05$ for the models over 1\,000 wild bootstrap (resp.\ permutations for the MMD) repetitions for various sample sizes $n$. 
Figure~\ref{fig:rotamers} shows the results. 
In particular, we observe in (a) that both tests achieve the nominal level (up to statistical fluctuations). (b) shows that both tests have a similar power under a global shift, while (c) shows that CKSD achieves higher power in the more challenging setting of a local shift. As (d) illustrates, the runtime of both methods is comparable, with that of CKSD being slightly larger.

\subsection{Permutations}
\label{sec:experiments:permutations}

Permutations are a common discrete data type with applications, among others, in label ranking~\citep{fotakis2022label}, in sports results analysis~\citep{dong2026ranking}, and in the analysis of experimental results~\citep{matteucci2023encoders}.
They often depend on covariates, such as, respectively, instances, competitors, or experimental factors.

\textbf{Formalization}\;
We consider $\m X = (-1, 1)$ and $\m Y = S_6$, the symmetric group on $\{1,\ldots,6\}$, whose elements are permutations $\sigma = (\sigma_1\dots\sigma_6)$, $\sigma_i$ being the rank of item~$i$. 
 
\textbf{Stein Kernel}\;
As $\m Y$ is discrete, we apply the Stein kernel described in Section~\ref{sec:example-applications:discrete} with $d=1$.
We order $S_6$ according to the Steinhaus-Johnson-Trotter ordering (Chapter 29; \citealt{bird2010algorithms}), which guarantees that two adjacent permutations---including first and last---have the ranks of consecutive items swapped.
For example, two consecutive terms might be $(1234\textbf{56})$ and $(1234\textbf{65})$.
We use the Gaussian kernel with $\gamma=1$ on $\m X$ and the Mallows kernel $\ell_{\nu}(\sigma, \sigma') \coloneqq e^{-\nu n_d(\sigma, \sigma')}$ on $\m Y$ \citep{jiao2015kendall}. In the latter, $n_d \coloneqq \tfrac14 \sum_{i, j =1}^{6} \lra{\sign\lrp{\sigma_i -\sigma_j} - \sign\lrp{\sigma'_i - \sigma'_j}}$ is the number of discordant pairs ($\sign(0) = 0$) and $\nu\coloneqq\binom62^{-1}$.

\textbf{Experimental Setup}\;
We set $\refdistrX = \m U(-1, 1)$ and assume that $\refdistrYx$ has pmf $p_x(\sigma) \propto \ell_{\lra{x}}(\sigma, c)$ for fixed $c=(123456) \in S_6$.
We compare three models $Q^i_{Y|\cdot}$ ($i=1, 2, 3)$, whose respective pmfs are
\begin{align}
    q^1_x(\sigma) &\propto \ell_{\lra{x}}(\sigma, c),    \tag*{$H_0$\quad} \\
    q^2_x(\sigma) &\propto \ell_{\lra{5x}}(\sigma, c), \tag*{$H_1$ (global shift)\quad} \\
    q^3_x(\sigma) &\propto 
        \begin{cases}
            q^1_x(\sigma) \quad \text{w.p.}\ 0.8,\\
            q^2_x(\sigma) \quad \text{w.p.}\ 0.2.
        \end{cases}
          \tag*{$H_1$ (local shift)\quad}
\end{align}
We use the same settings for level, bootstrap (resp.\ permutation) repetitions, and $n$ as in Section~\ref{sec:experiments:rotamers}.
Figure~\ref{fig:permutations} shows the results. 
While both CKSD and the MMD achieve the nominal level (up to sampling errors) in (a) and comparable power under a global shift in (b), CKSD obtains a higher power under a local shift in (c), which is barely detected by the MMD.
The runtimes of both approaches, displayed in (d), are similar.

\subsection*{AI use statement}

In this work, we used generative AI tools to help
implement the proposed method.
We have not used generative AI tools for synthetic-data generation, to help with theoretical models or conceptual developments, to prove mathematical claims, to assist in the writing of proofs, to propose or refine research methodology, to assist in translation, to clean and reformat datasets, to support qualitative data analysis, nor in the interpretation of results.
We have reviewed all AI-assisted work. 
LLM-generated code was verified and tested for correctness by the lead author.
We take responsibility for the final content of this work, including text, claims, or artifacts produced with the aid of generative AI.

\subsubsection*{Acknowledgements}
FK is funded by the KiKIT (The Pilot Program for Core-Informatics at the KIT) of the Helmholtz Association.

\bibliography{bib/unionbib,bib/publications_Florian}

\clearpage
\setcounter{equation}{0}
\renewcommand{\theequation}{\thesection.\arabic{equation}} %
\appendix
\thispagestyle{empty}

\onecolumn
 
\section{PROOFS}
\label{app:proofs}

This section is dedicated to the proofs of the results presented in the main text. The proof of Theorem~\ref{thm:mainresult} is in Appendix~\ref{sec:proof-mainresult} and that of Corollary~\ref{corr:cksd-is-characterizing} is in Appendix~\ref{sec:proof-cksd-is-characterizing}. We derive Lemma~\ref{lemma:estimator} in Appendix~\ref{sec:proof-estimator}, prove Lemma~\ref{lemma:convergence-rate} in Appendix~\ref{sec:proof-of-consistency}, and show Lemma~\ref{lemma:H0iffjointH0} in Appendix~\ref{app:proof_H0iffjointH0}.

\subsection{Proof of Theorem~\ref{thm:mainresult}} 
\label{sec:proof-mainresult}

We have shown in \eqref{eq:sufficient-Stein-condition} that if $\E_\refdistrYX[\psi_{\modelYX}(Y)] \overset{\refdistrX\aew}{=} 0_{\m H}$, then $\E_P[\Psi_Q(X,Y)] = 0_{\m G}$. To complete the proof, we show the converse in the following (with a similar idea to \citet[Proof of Theorem~1]{jitkrittum20testing} but weakened conditions). In other words, for the $\m X\to \m H$ function
\begin{align}
    h(x) \coloneqq \E_{\refdistrYx} [\psi_\modelYx(Y)] \in \m H,
\end{align}
we show that if $\E_P[\Psi_Q(X,Y)] = 0_{\m G}$, then $h(X) \overset{\refdistrX\aew}{=} 0_{\m H}$.

We start with an auxiliary observation. Let $L : L^2(\m X, P_X;\m H) \to L^2(\m X, P_X;\m H)$ be defined by
\begin{align}
    (Lf)(t) = \int_{\m X} K(t,x)f(x)\d P_X(x).
\end{align}
As $K$ is $C_0$-universal, $L$ is injective \citep[Theorem~2]{carmeli10vector} (recalled in Theorem~\ref{thm:carmeli-injective}). Using that $K(t,x) = \gamma_t^*\gamma_x$, and \citet[(A.32)]{steinwart08support} for exchanging the integral with the bounded linear operator, we get that
\begin{align}
    (Lf)(t) = \gamma^*_t\int_{\m X} \gamma_xf(x)\d P_X(x) \jover a= \gamma_t^*Sf,
\end{align}
where in (a) we define the linear operator $S : L^2(\m X, P_X;\m H) \to \m G$. We show that $S$ is injective.
Indeed, as $S$ is linear, $f=0_{L^2}$ implies $Sf = 0_{\m G}$.
In the other direction, by the linearity of $\gamma^*_t$, $Sf = 0_{\m G}$  implies $\gamma^*_t Sf = 0_{\m H}$; as this holds for every $t \in \m X$, we have $Lf = 0_{L^2}$ and, by injectivity of $L$, $f=0_{L^2}$.
Thus, $S$ is injective.

To prove the claim, we now observe that 
\begin{align}
    0_{\m G} = \E_P[\Psi_Q(X,Y)] \overset{\eqref{eq:exp-of-psi-Q}}{=} \E_{P_X}[\gamma_X\E_{\refdistrYX}\lrb{\psi_{\modelYX}(Y)}] = Sh,
\end{align}
which, by the injectivity  noted above, implies that $h(X) \overset{\refdistrX\aew}{=} 0_\m H$.

\subsection{Proof of Corollary~\ref{corr:cksd-is-characterizing}} 
\label{sec:proof-cksd-is-characterizing}
We have the chain of equivalences
\begin{align}
    D_Q(P) = 0 \jover a \iff \E_P[\Psi_Q(X,Y)] = 0_\m G \jover b \iff \E_\refdistrYX[\psi_{\modelYX}(Y)] \overset{\refdistrX\aew}{=} 0_{\m H} \jover c \iff  \modelYX \overset{\refdistrX\aew}{=} \refdistrYX,
\end{align}
with the following details. In (a), we use that $D_Q(P) = 0$ iff $D_Q^2(P) = 0$ and the equivalence
\begin{align}
     D_Q^2(P) &\overset{\eqref{eq:cksd-def}}{=} \norm{\E_P k_Q(\cdot,(X,Y))}{\m H_{k_Q}}^2 \jover d= \left\langle \E_P k_Q(\cdot,(X,Y)), \E_P k_Q(\cdot,(X,Y))\right\rangle_{\m H_{k_Q}}
     \jover e= \E_{P\otimes P}k_Q((X,Y),(X',Y'))  \\
    &\overset{\eqref{eq:def-k_Q}}{=}  \E_{P\otimes P} \left\langle \Psi_Q(X,Y), \Psi_Q(X',Y')\right\rangle_{\m G} \jover f=  \left\langle \E_P \Psi_Q(X,Y), \E_P \Psi_Q(X',Y')\right\rangle_{\m G} \jover g= \norm{\E_P[\Psi_Q(X,Y)]}{\m G}^2,
\end{align}
where (d) follows as in a Hilbert space the norm is induced by its inner product. In (e), we apply twice \citet[(A.32)]{steinwart08support} to interchange the integral with the inner product, and the reproducing property. (f) is by \citet[(A.32)]{steinwart08support}, and (g) as the inner product induces the norm. (b) follows by Theorem~\ref{thm:mainresult}. (c) holds by Assumption~\ref{ass:well-defined-RKHS}.
This proves the corollary.

\subsection{Proof of Lemma~\ref{lemma:estimator}} \label{sec:proof-estimator}

By the definition of CKSD in \eqref{eq:cksd-def} and substituting $P\mapsto \hat P_n$, we get
\begin{align}
    D_{Q}^2(\hat \refdistr_n) 
        &= \lrn{\E_{\hat \refdistr_n} k_Q\lrp{\cdot, (X, Y)}}_{\m H_{k_Q}}^2 = \E_{\hat \refdistr_n \otimes \hat \refdistr_n} \lrb{\lrang{k_Q\lrp{\cdot,(X, Y)}, k_Q\lrp{\cdot, (X', Y')}}_{\m H_{k_Q}}}\\
        &= \E_{\hat \refdistr_n \otimes \hat \refdistr_n}  [k_Q\lrp{(X, Y), (X', Y')}] = \frac{1}{n^2}\sum_{i, j = 1}^n k_Q\lrp{(X_i, Y_i), (X_j, Y_j)} = \frac{1}{n^2} \bm1_n^\transpose \b{K}_{n,n}\bm1_n,
\end{align}
which is the first part of the lemma. For the second part, we notice that the definition of $k_Q$ in \eqref{eq:def-k_Q} and the assumption of $K(x,x') = k(x,x')I_{\m H}$ imply that, for $x,x'\in \m X$ and $y,y'\in\m Y$,
\begin{align}
    k_Q((x,y),(x',y')) \overset{\eqref{eq:def-k_Q}}{=} \lrang{\psi_\modelYx(y), K(x, x') \psi_{Q_{Y|x'}}(y')}_{\m H} = k(x,x')\lrang{\psi_\modelYx(y), \psi_{Q_{Y|x'}}(y')}_{\m H}
\end{align}
by the linearity of the inner product. Hence, 
\begin{align}
    \b K_{n,n} &= [k_Q((X_i,Y_i),(X_j,Y_j))]_{i,j=1}^n = \left[k(X_i,X_j)\lrang{\psi_{Q_{Y\mid X_i}}(Y_i), \psi_{Q_{Y\mid X_j}}(Y_j)}_{\m H}\right]_{i,j=1}^n \\
    &= \left[k(X_i,X_j)\right]_{i,j=1}^n\circ\left[\lrang{\psi_{Q_{Y\mid X_i}}(Y_i), \psi_{Q_{Y\mid X_j}}(Y_j)}_{\m H}\right]_{i,j=1}^n = \b L_{n,n}\circ \b G_{n,n},
\end{align}
which concludes the derivation.

\subsection{Proof of Lemma~\ref{lemma:convergence-rate}} \label{sec:proof-of-consistency}

By \citet[Theorem~4]{kalinke26nystroemksdtests} (recalled in Theorem~\ref{thm:kalinke-psi1}), it suffices to show that 
\begin{align}
    \norm{\norm{k_Q(\cdot,(X,Y))}{\m H_{k_Q}}}{\psi_1} < \infty, \label{eq:sub-gauss-k_Q}
\end{align}
which follows from the chain of (in)equalities,
\begin{align}
    \MoveEqLeft\norm{\norm{k_Q(\cdot,(X,Y))}{\m H_{k_Q}}}{\psi_1} \jover a= \norm{\norm{\gamma_X\psi_{\modelYX}(Y)}{\m G}}{\psi_1} \jover{b}\le \norm{\norm{\gamma_X}{\text{op}}\norm{\psi_{\modelYX}(Y)}{\m H}}{\psi_1} \\
    &\jover c \le \norm{\sup_{x\in\m X}\norm{\gamma_x}{\text{op}}\norm{\psi_{\modelYX}(Y)}{\m H}}{\psi_1} \jover d\le M \norm{\norm{\psi_{\modelYX}(Y)}{\m H}}{\psi_1} \jover e< \infty,
\end{align}
with the following details. In (a), we use that $k_Q$ is feature map induced [\eqref{eq:def-k_Q}]. (b) is by a property of the operator norm and the monotonicity of the $\psi_1$-norm. In (c), we also use the monotonicity of the $\psi_1$-norm. (d) holds by using that for the $C_0$ kernel $K$, we have by \citet[Proposition~2]{carmeli10vector} (recalled in Theorem~\ref{thm:carmeli-uniform-bounded}) that there exists $M \ge 0$ such that $\sup_{x\in\m X}\norm{K(x,x)}{\text{op}} \le M^2$. Hence, using the $C^*$-property of the operator norm, we have, for any $x\in \m X$, $\norm{\gamma_x}{\text{op}}^2 = \norm{\gamma_x^*\gamma_x}{\text{op}} = \norm{K(x,x)}{\text{op}} \le M^2$. The finiteness in (e) now follows from $M< \infty$ and as $\norm{\norm{\psi_{\modelYX}(Y)}{\m H}}{\psi_1} <  \infty$ was assumed.

\subsection{Proof of Lemma~\ref{lemma:H0iffjointH0}}
\label{app:proof_H0iffjointH0}

For any $A \times B \in \Sigma_{\m X} \otimes \Sigma_{\m Y}$, we have
\begin{align}
    P(A \times B) &= \int_{A} \refdistrYx(B)  \d\refdistrX{(x)},\\
    Q(A \times B) &= \int_{A} \modelYx(B) \d\refdistrX{(x)}.
\end{align}
If $P = Q$, the l.h.s.-s are equivalent and therefore $\refdistrYX(B) \overset{\refdistrX\aew}= \modelYX(B)$. As this holds for any $B \in \Sigma_{\m Y}$, $\refdistrYX \overset{\refdistrX\aew}= \modelYX$.
Conversely, if $\refdistrYX(B) \overset{\refdistrX\aew}= \modelYX(B)$, the r.h.s.-s are equivalent and following the chain of equalities, we have that $P=Q$.

\section{EXTERNAL RESULTS}

This section collects the external results that we use; for brevity, we recall only the parts that we invoke.
\begin{theoremA}[Theorem 2; \citealt{carmeli10vector}] \label{thm:carmeli-injective}
    Let $(\m X, \tau_{\m X})$ be a locally compact second-countable topological space, $(\m H, \langle \cdot,\cdot\rangle_{\m H})$ a real separable Hilbert space, and $K: \m X \times \m X \to \m B(\m H, \m H)$ a $C_0$ $\m H$-kernel. 
    The following are equivalent.
    \begin{enumerate}[label=(\roman*)]
        \item $K$ is $C_0$-universal.
        \item For all $P \in \m M_1^+(\m X)$, the operator $L_P: L^2(\m X, P; \m H) \to L^2(\m X, P; \m H)$ given by
        \begin{align}
            (L_Pf)(x) = \int_{\m X}K(x,t)f(t)\d P(t)
        \end{align}
        is injective.
    \end{enumerate}
\end{theoremA}

\begin{theoremA}[Proposition 2; \citealt{carmeli10vector}] \label{thm:carmeli-uniform-bounded}
      Let $(\m X, \tau_{\m X})$ be a locally compact second-countable topological space, $(\m H, \langle \cdot,\cdot\rangle_{\m H})$ a real separable Hilbert space, and $K: \m X \times \m X \to \m B(\m H, \m H)$ an $\m H$-kernel.
    If $K$ is $C_0$ then the function $x \mapsto \lrn{K(x, x)}_{\mathrm{op}}$ is bounded for all $x\in \m X$.
\end{theoremA}

\begin{theoremA}[Theorem 4; \citealt{kalinke26nystroemksdtests}] \label{thm:kalinke-psi1} Let $(\m X,\tau_{\m X})$ be a separable topological space, $P\in \m M_1^+(\m X)$, $X_1,\ldots,X_n\iid P$, $k : \m X \times \m X \to \R$ a continuous kernel, and $\E_{X\sim P}\sqrt{k(X,X)} < \infty $. If $\norm{\norm{k(\cdot,X)}{\m H_{k}}}{\psi_1} < \infty$, then 
	\begin{align}
		\big\|\mu_k\big(P\big)\big\|_{\m H_k} -\big\|\mu_k\big(\hat P_n\big)\big\|_{\m H_k} = O_P\big(n^{-1/2}\big).
	\end{align}
\end{theoremA}

\end{document}